# Physics-Guided Hierarchical Reward Mechanism for Learning-Based Robotic Grasping

Yunsik Jung*, Lingfeng Tao*, Michael Bowman**, Jiucai Zhang^,
and Xiaoli Zhang*, *Member*, IEEE

*Abstract*—Learning-based grasping can afford real-time grasp motion planning of multi-fingered robotics hands thanks to its high computational efficiency. However, learning-based methods are required to explore large search spaces during the learning process. The search space causes low learning efficiency, which has been the main barrier to its practical adoption. In addition, the trained policy lacks a generalizable outcome unless objects are identical to the trained objects. In this work, we develop a novel Physics-Guided Deep Reinforcement Learning with a Hierarchical Reward Mechanism to improve learning efficiency and generalizability for learning-based autonomous grasping. Unlike conventional observation-based grasp learning, physics-informed metrics are utilized to convey correlations between features associated with hand structures and objects to improve learning efficiency and outcomes. Further, the hierarchical reward mechanism enables the robot to learn prioritized components of the grasping tasks. Our method is validated in robotic grasping tasks with a 3-finger MICO robot arm. The results show that our method outperformed the standard Deep Reinforcement Learning methods in various robotic grasping tasks.

## I. Introduction

Multi-finger robotic grasping is critical to accomplish object manipulation that can replace human activities in various environments, such as the manufacturing industry, space, and deep-sea maintenance [1, 2]. Despite the potential, it is still challenging in several aspects. High-dimensional robotic hands capable of performing complex tasks have been developed. However, the inherent high dimensionality of these hands and their interactions with objects often result in a large search space, leading to low learning efficiency and a significant data requirement for training purposes. In addition, interactions between robotic hands and objects with various contours are demanding to accomplish stable performances. Therefore, the adoption of multi-finger robotic grasping is limited [4].

The Deep Reinforcement Learning (DRL) method has significantly improved performance in robotic grasping by handling high-dimensional problems and enabling real-time autonomous grasping [8]. It provides an optimal control policy that maximizes an objective function/reward based on the specific state of an autonomous grasping task. To address the issue of high-dimensionality in robotic grasping, various approaches have been employed, such as learning from human demonstrations, dimensionality reduction using subspaces, and incorporating vision-based support [3, 15].

However, a common issue of DRL approaches and other observation-based robot learning is that training a robot is time-consuming to reach sufficient stabilities and performances. Training requires exploring a broad search space because of complex configurations of robotic hands and interactions with objects, which results in low learning *efficiency*. Further, the *generalizability* of the trained policy is limited unless they are identical to the trained objects. One of the critical reasons for the learning efficiency and the generalizability issues is that current DRL methods mainly use task-related sparse reward components as the only criterion to define the reward function [9]. However, a successful grasp usually requires multiple quality-related criteria components, such as grasp pose, contact points/regions on the object, and grasp stability. These quality-related criteria are commonly used in physics-based grasping methods but have rarely been considered in learning-based grasping. Thus, it is not easy for current learning-based robots to fundamentally *understand* how to achieve and improve grasp quality other than task completion. Ideally, considering both task completion and these quality-related criteria as the reward can help RL-based robots to explore the environment and generalize the learned policy efficiently.

This paper introduces the physics-guided DRL with a hierarchical reward mechanism (PG-H-RL) for autonomous grasping and leverages both the positives of learning and physics to facilitate computationally efficient yet high-quality grasping solutions by enabling the robot to understand the problem at hand fundamentally. The contributions of this work are:

- Developed a physics-guided learning strategy for autonomous grasping, by integrating physics-based metrics as rewards to guide the robot to understand the grasping task. This improved learning efficiency and yielded physically consistent performances.
- Developed a hierarchical reward mechanism to learn the task prioritized physics-based rewards which allows the robot to understand different metrics and improves learning efficiency.
- Validated the generalizability of the trained policy of PG-H-RL on the Mico arm using objects of different sizes compared to the object used during training.

*Y. Jung, L. Tao, and X. Zhang are with Colorado School of Mines, Intelligent Robotics and System Lab, Golden, CO 80401 USA (e-mail: yunsikjung@mines.edu, tao@mines.edu, mibowman@mines.edu) (phone: 303-384-2343; fax: 303-273-3602; e-mail: xlzhang@mines.edu)

**M. Bowman is with the Cancer Biology Department of University of Pennsylvania, the Bowman Lab, Philadelphia, PA 19104 USA (e-mail: Michael.Bowman@Pennmedicine.upenn.edu)

^J. Zhang is with the GAC R&D Center Silicon Valley, Sunnyvale, CA 94085 USA (e-mail: zhangjiucai@gmail.com)

Acknowledgement: This material is based on work supported by the US NSF under grant 1652454 and 2114464. Any opinions, findings, conclusions, or recommendations expressed are those of the authors and do not necessarily reflect those of the National Science Foundation.

## II. RELATED WORK

### A. Analytical Methods for Autonomous Grasping

Analytical approaches consider physics, kinematics, and dynamics of objects and hands to get the correct grasp, which is a vital aspect of accomplishing grasping tasks. In [13], they proposed a grasping simulator to compute robot and object motions under the influence of external forces and contacts. Form closure and force closure properties are basic grasp quality criteria used in these approaches to find the grasp [11]. Grasp quality measures have been developed to interpret the quality of robotic grasping. The measures associated with contact points on the object and the configurations of the robotic hands were developed [12, 13]. In [5], they introduced an approach combining empirical and analytical methods by imitating humans to reduce the computation time of calculating force-closure grasps. Finding the optimal solution to meet these criteria requires high computational power and limits real-world applications.

### B. Learning-based Methods for Autonomous Grasping

DRL for robotic grasping has been actively studied in recent years. Earlier work aimed to acquire strategies for robotic grasps using DRL with images [6]. Recently, DRL methods have been proposed to accomplish autonomous grasping with vision-based observations [7]. However, these learning methods require extensive number of sensors, training data and time to explore. To overcome this, human preferences, demonstration data, and potential contact regions were utilized [14, 15, 16]. In [17], they estimated the probability of a successful grasp using the contact region database collected from human demonstrations. The probability was considered as a partial reward to increase learning speed. Although these pure learning approaches could improve learning efficiency by reducing the search space, they do not fundamentally learn physics as a physics-informed approach would.

### C. Physics-guided Learning Strategies

Despite the validated effectiveness of physics-guided learning in other applications [18], few studies [16, 20, 21] have been reported in the robotic grasping field. In [19], they used the physics-guided target poses as the input for the learning process to improve performance for manipulation tasks on a physics simulator. In [16], they proposed an RL method that utilized a grasp quality metric as a binary reward. A good grasping configuration was determined by comparing potential grasp locations, contact information, and object from a database of successful grasp attempts. In addition, [20] defined the reward function that summed the force-closure quality index [21]. However, all these methods treated the grasp quality as binary bonus rewards and used a linear summation which can be easily biased or lose the grasping information.

### D. Hierarchical Reward in RL

The reward formulation of DRL is usually a linear summation of the reward components [9, 16, 17], which is implicit and inefficient for learning the multi-objective priorities and causes poor learning performance for multi-objective tasks (i.e., it takes a long time to learn or even fails to learn a correct policy). Hierarchical reward methods have been proposed to enable a robot to learn multi-objective tasks such as achieving autonomy or human-like merging actions for driving [22] and performing home service activities [23]. The formulation of the reward hierarchies contains logical or weighted connections. Logical connections are strict constraints, where the higher-level hierarchy must be learned before the lower-level one. Weighted connections are soft constraints, where the higher-level and lower-level hierarchies are learned together with a weighted summation. In [24], an RL agent for swarm robot control is trained with a logically connected hierarchical reward function. Inspired by these studies, this paper introduces the hierarchical reward in physics-guided grasp learning to learn multiple physics metrics and their correlations explicitly and efficiently.

## III. METHODOLOGY

PG-H-RL is developed to enable a multi-finger robot to learn grasping and lifting an object to a target height, and perform additional practical tasks: basic grasping, relocation, alignment tasks. It is assumed that the object's position and contact information between the robotic hand and the object can be detected to calculate the grasp quality.

### A. Reinforcement Learning Formulation

The grasping task is formulated as an RL problem that follows the Markov Decision Process (MDP). The MDP is defined as a tuple $\{S, A, R, \gamma\}$, where $S$ is the state of the environment and $A$ is the set of actions. $R(s'|s, a)$ is the reward function to give the reward after the transition from state $s$ to state $s'$ with action $a$ and $\gamma$ is a discount factor. Since the task is required to consider interactions with objects and more accurate controls, the continuous control domain is considered. To solve the problem, PG-H-RL adopts the Twin Delayed Deep Deterministic policy gradient (TD3) algorithm, which is a model-free reinforcement learning [25].

### B. Physics Metrics and Constraints

*1) Contact points interacting with an object*

Grasp quality is important for the agent to achieve a stable grasp. Contacts between the robotic hand and the object are important and necessary to evaluate grasp quality. The grasp matrix $G$ is defined by the relevant velocity kinematics and force transmission properties of the contacts on convex and non-convex objects in three-dimensional space [13]. In this work, two measures of grasp quality are computed with $G$: the measure for being graspable ($r_{graspable}$) and the normalized volume of the ellipsoid ($r_{vew}$). Inspired by previous studies with the traditional criteria of physics metrics as binary evaluations, the null space of the grasping matrix $\mathbb{N}(G)$ is considered a reward to indicate whether a grasp is graspable or ungraspable based on internal object forces [13].

$$r_{graspable} = \begin{cases} 0 & \mathbb{N}(G)=0 \\ 0.1 & \mathbb{N}(G)\neq 0 \end{cases} \quad (1)$$

where $\mathbb{N}(G)\neq 0$ reveals being graspable. It judges the grasp quality by providing an initial guide before further evaluation, which can reduce the search space. The binary value is empirically determined considering its importance level relative to other components in the entire reward. Using $G$, the contribution of all contact forces on the object can

refer to the continuous grasp quality measure using the volume of the ellipsoid in the wrench space [15] as:

$$Q_{vew} = \sqrt{\det(GG^T)} = \sigma_1 \sigma_2 \sigma_3 \cdots \sigma_m \quad (2)$$

where, $\sigma_1, \sigma_2, ..., \sigma_m$ denotes the singular values of $G$, and $m$ is the number of contact points on the object. This value is continuous and must be maximized to obtain the optimum grasp. The maximum value of $Q_{vew}$ is affected by the number of contacts, so the reward, $r_{vew}$, is the normalized value as:

$$r_{vew} = \text{norm}(Q_{vew}) \quad (3)$$

Fig. 1 (a) illustrates the examples of an optimal grasp of $Q_{vew}$. An example of robotic grasps with their $r_{vew}$s is in Fig. 1 (b).

  *2) Hand pre-shaping constraints*

Constraints were placed on the fingers to *prevent closing the fingers* and *avoid contact between the fingers*. These constraints act as penalties to the reward to pre-shape the robot hand for the grasping tasks. The former provides a penalty when the fingers attempt to close before the hand approaches the pre-grasp distance, and the latter provides a penalty if there is any contact between the fingers.

## C. Hierarchical Reward Mechanism with Physics Metrics

Using physics metrics, multiple components in the reward function are prioritized logically to learn autonomous grasping progressively. Firstly, an autonomous grasp task is broken down into three sequential stages: 1) approaching, 2) grasping, and 3) lifting. In the approaching stage, pre-shaping constraints are incorporated into the reward function prior to touching the object. These constraints are implemented as *penalties* within the reward function, facilitating proximity to the object, finger opening during this stage, and preventing finger-to-finger contact. These considerations contribute to achieving a successful grasp. The grasping stage includes grasping quality metrics with a hierarchical structure considering their priorities. The hierarchical structure reflects that a measure and/or constraint with a lower priority is not considered when a condition of one with a higher priority is not satisfied. It improves learning efficiency because the agent can explore action/state spaces efficiently depending on the satisfaction of the higher level hierarchies. During the lifting stage, a final task-related goal is considered, typically involving the error between the height of the object and the target height. The reward, $r_{obj\_height}$, provides a positive reward based on amount of the error to reach to the target height. The entire structure of the reward function and the hierarchical physics-guided reward mechanism are described in Fig. 2.

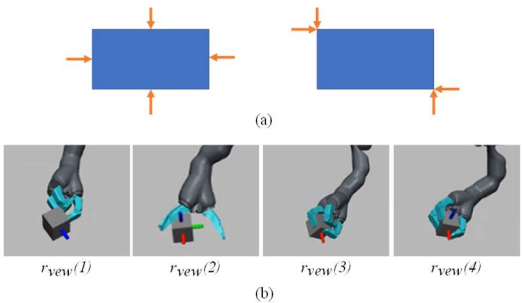

Figure. 1 The examples of the volume of ellipsoid in wrench space of $G$. (a) the optimal grasps with symmetric locations of contact points on the 2-D object. (b) grasps of the robotic hand. The grasp quality comparisons of these 4 $r_{vew}$s is $r_{vew}(1) < r_{vew}(2) < r_{vew}(3) < r_{vew}(4)$.

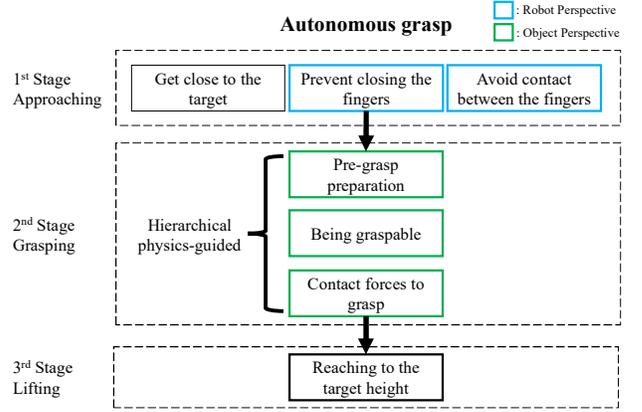

Figure. 2 The autonomous robotic grasp task is decomposed into three stages: approaching, grasping, and lifting. In the grasping stage, the hierarchical physics-guided mechanism is implemented. Blue lines and green lines represent robot and object perspectives, respectively.

In the grasping stage, *pre-grasp preparation* is a condition to determine whether the agent gets close enough to the object. To assign greater importance when the distance to the object is closer, the normalized exponential value of the distance between the robotic hand and the object ($r_{dist}$) is used as:

$$r_{dist} = \text{norm}\left(e^{0.1 \times dist_{obj\_hand}^{-1}}\right) \quad (4)$$

where $dist_{obj\_hand}$ is the distance between the hand and the object. *Being graspable* is a condition to be determined whether a grasp can be possible or not based on (1). *Contact forces to grasp* is a condition to show how much contact forces are applied to grasp the object based on $r_{vew}$ in (3). Fig. 3 shows the total reward function including each stage. In the grasping stage, $\lambda$, $\mu$, and $\nu$ are binary coefficients defined as 1 when both conditions of higher hierarchies and the corresponding condition are satisfied; otherwise, they are 0. They are defined as:

$$\lambda = \begin{cases} 0 & \text{if } dist_{obj\_hand} \geq d_{grasp} \\ 1 & \text{if } dist_{obj\_hand} < d_{grasp} \end{cases} \quad (5)$$

$$\mu = \begin{cases} 0 & \text{if } \mathbb{N}(G)=0 \\ \lambda \times 1 & \text{if } \mathbb{N}(G) \neq 0 \end{cases} \quad (6)$$

$$\nu = \begin{cases} 0 & \text{if } Q_{vew} \leq 0 \\ \lambda \times \mu \times 1 & \text{if } Q_{vew} > 0 \end{cases} \quad (7)$$

where $d_{grasp}$ is determined as 0.05m, considering the object size, to indicate entering the grasping stage.

## IV. EXPERIMENTS

### A. Experimental Setup

  *1) MICO arm:* To accomplish the task, a Kinova MICO

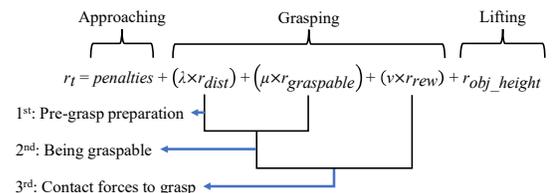

Figure. 3 The total reward function including all three stages. In the Grasping stage, the hierarchy is shown with the 1st, 2nd, and 3rd conditions.

arm is used [26], which has 6 rotational joints for the arm and the three-fingered robot hand. The MICO arm can be controlled to open and close the fingers to grasp the object.

*2) Simulator:* Mujoco Haptix is used as the simulator [27] since it provides high speed and accurate modeling for robotics, especially simulated hands grasping objects.

For the TD3 agent, we define the state observation, including the object's positions, the finger joint angles, the robotic hand's position, and the arm's joint angles. Table I shows the hyper-parameters for the TD3 agent. The object with a side length of 0.065m was used to train the policy. Three different tasks, namely basic grasping, relocation, and alignment tasks are designed to evaluate PG-H-RL. In the basic grasping task, it is aimed to grasp and lift the object to the target height with various sizes of the cube as ±5% to evaluate the generalizability of the policy. The object is randomly generated within the workspace of the robot. For the relocation task, the objective is to move the object to a specified target position. To improve the efficiency of the learning process, the trained policies for the basic grasping task were used as initial policies. Both the initial position and the target position are generated randomly.

Additionally, the alignment task is designed with the goal of rotating the object to a target orientation around the Z-axis, specifically within a range of 90 to 180 degrees. The purpose of this task is to ensure that the object is aligned correctly according to its orientation. The PG-H-RL for this task exhibits a lower degree of hierarchy in the reward structure, resulting in a reduced influence of the hierarchical reward mechanism. Therefore, PG-H-RL and Task Only were evaluated. The initial orientation and target orientation are randomly chosen.

These tasks are specifically designed as they are common and crucial for evaluating the stability of the trained policy in grasping. Unstable grasps have a significant risk of task failure, as they can result in the object being dropped or sliding during rotation or transfer. This highlights the sensitivity of task success to the quality of the grasp.

### B. Evaluation Methods and Metrics

Two different reward functions are considered baselines to evaluate the influence of the physics metrics and the hierarchical reward mechanism of the PG-H-RL method. Except for the reward functions, the baselines are in the same conditions and set up with PG-H-RL. A baseline, *Task Only*, has a reward function with task-related reward components:

$$r_{TaskOnly} = p_{target} + r_{obj\_height} \qquad (10)$$

where $p_{target}$ is a penalty that is determined based on $dist_{obj\_hand}$. The second baseline, Linear Summed, uses all reward components considered in PG-H-RL, but a linear summation is used instead of a hierarchical reward mechanism. The PG-H-RL method and the two baselines were trained with the same grasping task to evaluate the learning efficiencies and outcomes. Trends of the total reward with multiple training processes are used to compare the learning efficiency. The trends are evaluated by comparing how fast the total reward increases and maintaining this increase. To evaluate the learning outcomes, an error of height between the object's actual height and the target of 0.05m from the table is used. $Q_{vew}$ is considered to assess the stable and firm grasp quality. The distance to the object center is the distance between the desired location of the object center and the actual object center to evaluate the stability of a grasp pose (in Fig. 6 (a)). Further, the dropping rate of the object is analyzed to demonstrate the impact of grasp quality. The success of the basic grasping task is evaluated based on the height error, which should be less than 0.05m. For the relocation task, success is determined by the object's location error, which should be within 0.10m. In the alignment task, success is determined by the orientation error, which should be less than 10 degrees.

## V. RESULTS AND DISCUSSION

### A. Hierarchical Reward Mechanism

Fig. 4 shows an episode reward using PG-H-RL in late training. The duration from 0 to 21 steps indicates the robotic hand is approaching the object. Then, the reward entered the next stage for grasping with three physics-guided reward components associated with the object's perspective. The *pre-grasp preparation* condition is the first hierarchy and is satisfied after 21 steps. The *being graspable* condition is the second hierarchy and is satisfied after 25 steps. The *contact forces to grasp* condition is the third hierarchy and is satisfied after 29 steps. The hierarchical reward mechanism allows the agent to learn higher level constraints before lower ones. In the following section B and C, a detailed comparison between PG-H-RL and Linear Summed in terms of success rates and performances are presented. The results clearly demonstrate that by considering relative priorities of multiple components in the learning process, PG-H-RL significantly enhances learning processes.

### B. Learning Efficiency

The results of the training processes are presented in Fig. 5, illustrating the performance across all tasks. It is observed that the relocation task and the alignment task pose greater difficulty compared to the basic grasping task. While both tasks necessitate reaching the target location or orientation, the basic grasping task involves lifting the object from the table. The multiple training processes exhibit consistent learning curves for all methods. Notably, PG-H-RL demonstrates the highest learning efficiency across all tasks within the same number of training episodes. The learning speeds of PG-H-RL and Linear Summed are faster than Task Only. This validates that the physics metrics and constraints as the reward effectively guides agents to learn the tasks faster. Further, a comparison between the results of PG-H-RL and Linear Summed reveals that introducing multiple objectives without considering their relative priorities confuses the agent in balancing among different reward components and achieving high rewards within a certain number of training episode.

TABLE I. HYPER-PARAMETERS FOR TD3 ALGORITHM

| Parameters | Values |
|---|---|
| Sample Time | 0.05 [sec] |
| Discount Factor | 0.99 |
| Mini-Batch Size | 256 |
| Experience Buffer Length | 1e+6 |
| Target Smooth Factor | 0.005 |
| Learning rate | 0.001 |
| Target Update Frequency | 2 |
| Sequence Length | 1 |

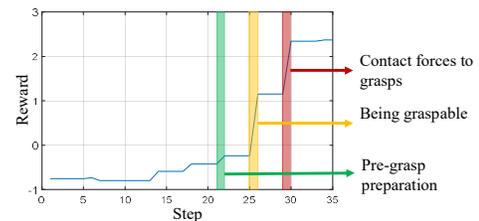

Figure. 4 The reward evolution for a single episode using PG-H-RL.

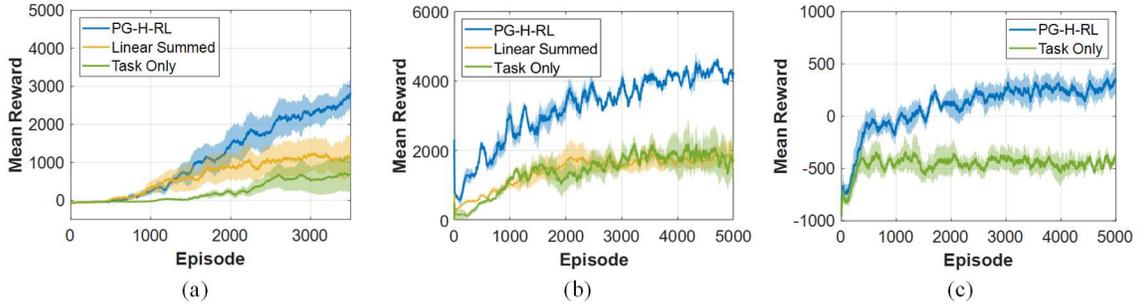

Figure. 5 The learning curves with mean total rewards for (a) the basic grasping, (b) the relocation, and (c) the alignment tasks. The shaded areas indicate the standard deviations for the multiple training processes.

### C. Learning Outcome

Table II presents the success rates achieved in the basic grasping task with multiple training processes. It can be observed that PG-H-RL consistently outperforms both the Linear Summed and Task Only approaches across different object sizes. These results indicate that the integration of physics-informed metrics and the hierarchical reward mechanism not only enhances the learning efficiency but also significantly improves the overall success rate in performing the task. Additionally, the success rates achieved by PG-H-RL across various sizes of objects remain consistently higher, indicating that the trained policy is generalizable to similar objects as the one used for the training. Further, Fig. 6 (c) and (d) illustrate that PG-H-RL achieves closer distances to the object and the higher the grasp quality, which are essential to stable grasps. As a result, PG-H-RL achieves the highest success rates for this task compared to the others. This is also observed with the

TABLE II. RESULTS OF SUCCESS RATES AND PERFORMANCE FOR THE BASIC GRASPING TASK

| Size | PG-H-RL | Linear Summed | Task Only |
|---|---|---|---|
| -5% | 45 | 25 | 27 |
| Original | 63 | 20 | 30 |
| +5% | 54 | 28 | 28 |

a. The values of the success rates are percentages.

TABLE III. MESURES FOR PERFORMANCE EVALUATION

| Metric | PG-H-RL | Linear Summed | Task Only |
|---|---|---|---|
| Error of Height [m] | 0.0726 | 0.0449 | 0.0414 |
| Distance to the Object Center [m] | 0.0913 | 0.1253 | 0.1846 |
| Grasp Quality [$Q_{vew}$] | 27.8171 | 9.7428 | 10.3109 |

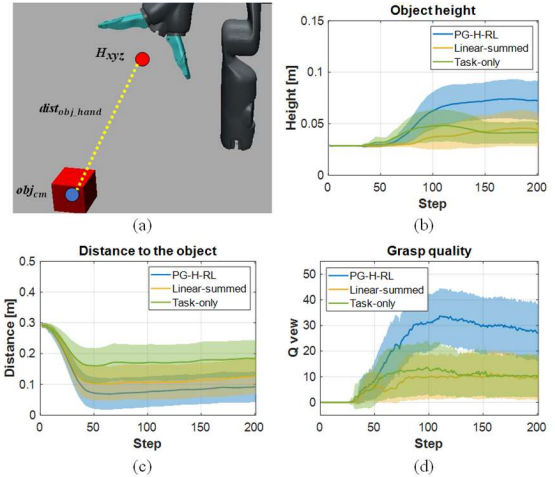

Figure. 6 (a) illustrates the parameters, $H_{xyz}$ is end-effector of the robotic hand where the object center is desired to locate, $Obj_{cm}$ the center of the object, $dist_{obj\_hand}$ is a distance to the object center. The performance comparisons are done using three measures: Object height (b), distance to the object center (c), and grasp quality (d).

object height in Fig. 6 (b). Table III shows the average values for the evaluation metrics across the multiple evaluation trials, aligning with the findings depicted in Fig. 6. The results provide compelling evidence for the effectiveness of the proposed PG-H-RL, highlighting its superiority over alternative approaches in achieving

TABLE IV. P-VALUES USING ONE-WAY ANALYSIS OF VARIANCE FOR LEARNING PERFORMANCES

| Category | PG-H-RL vs Linear Summed | PG-H-RL vs Task Only |
|---|---|---|
| Object Height | 4.3311e-29 | 1.561e-20 |
| Distance to the Object Center | 2.8393e-06 | 3.2394e-48 |
| Grasp Quality | 8.2321e-43 | 1.8412e-35 |

a. The One-way Analysis of variance for the performance data can be obtained using the MATLAB function (ANOVA1).

TABLE V. P-VALUES OF N-1 CHI-SQUARE TEST FOR SUCCESS RATES

| Size | PG-H-RL vs Linear Summed | PG-H-RL vs Task Only |
|---|---|---|
| -5% | 0.0031 | 0.0082 |
| Original | 7.4848e-10 | 3.05771e-13 |
| +5% | 0.0018 | 0.0018 |

a. $x^2$ can be calculated using the binary result (pass and fail) and the total number of trials [25].
b. The N-1 Chi-squared distribution (p-value) can be obtained using $x^2$ and a table of Chi-square values or the Excel function (CHIDIST).

TABLE VI. BEST PERFORMANCE FOR ALL THE TASKS

| | Basic grasping | | | Relocation | | | Alignment | | |
|---|---|---|---|---|---|---|---|---|---|
| | Success rate | Object height | Grasp quality | Success rate | Mean error | Grasp quality | Success rate | Mean error | Grasp quality |
| PG-H-RL | **91** | **0.0911** [0.0257] | **29.3330** [9.3449] | **44** | **0.1324** [0.0916] | **27.5305** [10.6220] | **45** | **25.0995** [24.2733] | **11.8740** [2.6859] |
| Linear Summed | 76 | 0.0987 [0.0421] | 25.5575 [10.9779] | 27 | 0.1551 [0.0826] | 15.6718 [11.6193] | N/A | N/A | N/A |
| Task Only | 57 | 0.0570 [0.0250] | 18.9022 [14.9108] | 29 | 0.1691 [0.1081] | 15.3582 [13.3574] | 15 | 74.1100 [47.7846] | 4.8435 [4.0635] |

a. Success rate [%], Object height [standard deviation] [m], Grasp quality [standard deviation] [$Q_{vew}$]

successful grasps. These findings emphasize the importance of incorporating physics-based guidance and the hierarchical reward structure in the robotic grasping task. To determine statistically significant differences in the comparisons of the evaluation metrics, we conducted a one-way analysis of variance (ANOVA) [29] (Table IV). Additionally, for the success rates, we performed the N-1 Chi-Square test [28] (Table V). A significance level of $p < 0.05$ was used to indicate statistically significant differences.

Table VI illustrates the best performances for each task, including success rates and relevant evaluation metrics to assess task performances. It is evident that PG-H-RL consistently outperforms the Linear Summed and Task Only approaches in all the tasks. Comparing the success rates achieved in the basic grasping task with multiple training processes, it is evident that even the best performances of the Linear Summed and Task Only approaches achieved higher success rates, the PG-H-RL approach outperformed them with the highest success rate. Although the Linear Summed approach achieved a higher object height, its significantly greater standard deviation suggests instability. On the other hand, PG-H-RL consistently outperforms other methods in terms of grasp quality, which aligns with the results obtained from multiple training processes and supports the interpretation that PG-H-RL is more effective and reliable in achieving superior grasp quality (Fig. 7 (a)). Regarding the results of the relocation task, the models trained with the Task Only approach within a certain number of training episodes demonstrated the ability to approach the object, but were unable to successfully grasp and relocate it to the target location. Additionally, the models trained with the Linear Summed approach achieved some successes in the relocation task. However, the success rate was significantly lower compared to the performance of the PG-H-RL approach. The average errors between the object and the target location further highlights the superiority of the PG-H-RL. It achieved lower error value compared to the other approaches. Fig. 7 (b) illustrates the grasp quality achieved in the relocation task over different steps. It can be observed that PG-H-RL consistently achieves more secure grasps with the object compared to Linear Summed. This highlights the benefits of the hierarchical reward mechanism in learning multiple reward components. The results achieved in the alignment task, the consistent aspect about the benefits of PG-H-RL is observed. The model trained with the Task Only showed some successful realignments of the object, however, it shows a significantly lower success rate than PG-H-RL. Since the initial difference is set between 90 to 180 degrees, there is a noticeable gap in the average errors between the PG-H-RL and Task Only approaches. It highlights the superior performance of PG-H-RL in achieving accurate alignments with the target orientation. Fig. 7 (c) illustrates the grasp quality achieved in the alignment task over different steps. It is evident that PG-H-RL consistently achieves higher grasp qualities compared to Task Only. This demonstrates the effectiveness of the physics-informed metrics in guiding the model to achieve secure grasps, which ultimately leads to higher success rates in the alignment task. Unlike the relocation task, the alignment task exhibits a decrease in grasp quality after reaching the target orientation, as stable grasping is no longer required at that stage. Additional details and visual demonstrations of the performances can be found in the supplementary video.

## VI. CONCLUSION

This work first utilizes the physics informed metrics and the hierarchical reward mechanism to solve the tasks in the field of robotic grasping. The experimental results obtained from the proposed tasks clearly highlight the substantial advantages of employing these techniques. The use of physics-informed metrics and the hierarchical reward mechanism in the PG-H-RL framework has led to improved outcomes in terms of learning efficiency and overall task performance. These findings provide strong evidence for the effectiveness of these techniques in enhancing learning processes and achieving superior performance across diverse task scenarios. The promising outcomes of this research highlight the need for further exploration and application of these approaches in the field. Future studies can focus on refining and extending our method to address more complex grasping scenarios involving objects with intricate shapes. Additionally, further research can explore the feasibility of applying this method in real-world applications.

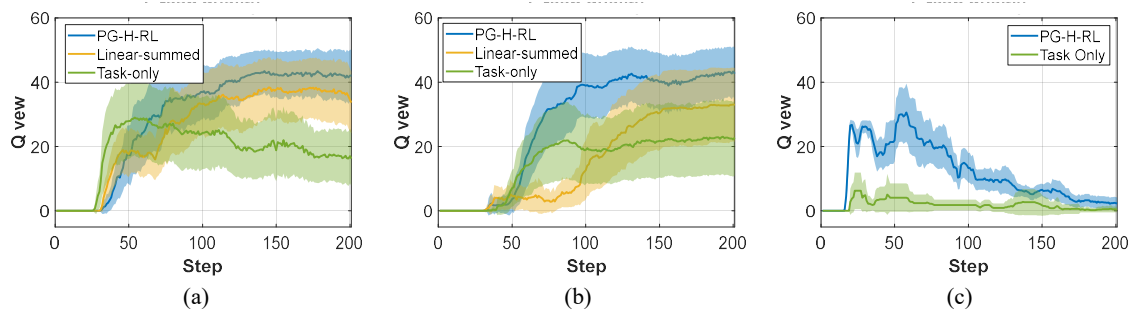

Figure. 7 The best performances of grasp quality. (a) is a grasp quality of the basic grasping task, (b) is a grasp quality of the relocation task, (b) is a grasp quality of the alignment task.